\documentclass[runningheads]{llncs}
\usepackage{graphicx}
\usepackage{amsmath,amssymb}
\usepackage{comment}
\usepackage{subfig}
\usepackage{cite}
\usepackage{relsize}
\usepackage{array}
\newcolumntype{P}[1]{>{\centering\arraybackslash}m{#1}}

\begin{document}
\title{Spiking-GAN: A Spiking Generative Adversarial Network Using Time-To-First-Spike Coding}
\author{Vineet Kotariya\and
Udayan Ganguly}
\authorrunning{V. Kotariya and U. Ganguly}
\titlerunning{Spiking-GAN: A Spiking Generative Adversarial Network}
\institute{Department of Electrical Engineering, IIT-Bombay, Mumbai, India
\email{vineetkotariya@iitb.ac.in}, \email{udayan@ee.iitb.ac.in}}
\maketitle
\begin{abstract}
Spiking Neural Networks (SNNs) have shown great potential in solving deep learning problems in an energy-efficient manner. However, they are still limited to simple classification tasks. In this paper, we propose Spiking-GAN, the first spike-based Generative Adversarial Network (GAN). It employs a kind of temporal coding scheme called time-to-first-spike coding. We train it using approximate backpropagation in the temporal domain.  We use simple integrate-and-fire (IF) neurons with very high refractory period for our network which ensures a maximum of one spike per neuron. This makes the model much sparser than a spike rate-based system. Our modified temporal loss function called `Aggressive TTFS' improves the inference time of the network by over 33\% and reduces the number of spikes in the network by more than 11\% compared to previous works. Our experiments show that on training the network on the MNIST dataset using this approach, we can generate high quality samples. Thereby demonstrating the potential of this framework for solving such problems in the spiking domain.
\end{abstract}
\section{Introduction}

Deep Neural Networks (DNNs) have significantly outperformed traditional algorithms and have set new performance benchmarks in a plethora of applications. However, the increase in complexity driven by its success has led to a keen interest in low-power deep learning techniques, especially for mobile and embedded applications.~\cite{surv_low_power}\\
\indent Spiking Neural Networks (SNNs) are third generation neural networks where binary `spikes' are the tokens of information. These networks are promising due to their power efficiency. SNNs are biologically plausible and attempt to mimic the neuronal dynamics of the brain. Information is coded in the form of spikes (inspired by action potentials in the brain). Implementation of these networks on event-driven neuromorphic hardware~\cite{loihi, truenorth, spinnaker} has been found to be both fast and energy efficient~\cite{spiking-yolo, low-power}. However, SNNs are notoriously difficult to train. This is primarily because the SNN neurons (activation functions) are non-differentiable which makes it hard to propagate the error through the network~\cite{DL_SNN,SNN_surv}. As a result most SNN applications have been limited to simple classification tasks~\cite{DL_SNN, illing2019}. There is a need to attempt a wider range of problems and explore more challenging tasks using SNNs.\\
\indent Generative Adversarial Networks (GANs)~\cite{GAN} are currently one of the most promising and extensively researched deep learning topics~\cite{GAN_surv}. GANs primarily consist of two networks, the generator and the discriminator, which compete against each other. 
The generator is trained to try and deceive the discriminator by generating synthetic/fake samples from a noise prior. The discriminator in turn tries to distinguish the generator’s samples from the real data by classifying them as fake or real. GANs have found an increasing number of applications~\cite{GAN_surv} like image generation, super-resolution, de-occlusion, image-to-image and text-to-image translation and even drug discovery, among others. However, there is no equivalent of a GAN in the spiking domain.
\\ \indent
Most SNNs have predominantly employed spiking rate-based coding schemes. Temporal coding provides an alternate framework. Time-to-first-spike (TTFS) coding is a type of temporal coding in which the information is encoded in the spike time of a neuron. It significantly increases the sparsity of the output spike train, thereby giving large energy savings. It has recently been used in SNNs for classification problems~\cite{s4nn, t2fsnn, eth18} and hardware implementation of such schemes is also being explored~\cite{ttfs_hw}. There is also building evidence of the biological plausibility of such temporal coding schemes~\cite{bio1,bio2}. TTFS coding provides benefits like increased sparsity and lower inference latency of the network over traditional rate-based methods.\\
\indent In this paper, we propose and demonstrate for the first time a TTFS-coding based spiking implementation of a simple Generative Adversarial Network called Spiking-GAN. We demonstrate the successful generation of good quality images which look natural while improving the sparsity of representation and training.
\section{Methods}
We have adopted the learning rule and neuron model (sec. 2.2, 2.3 and 2.5) described in S4NN~\cite{s4nn} and modified it for our network. 
\subsection{Spiking Neuronal Dynamics}
We have used a simple Integrate-and-Fire (IF) neuron model having a refractory period ($t_{ref}$) for our network. The membrane potential of the $j^{th}$ neuron in the $l^{th}$ layer at time t is given by:
\begin{figure}[htp]
\centering
    \subfloat[\centering ]{{\includegraphics[width=0.45\textwidth]{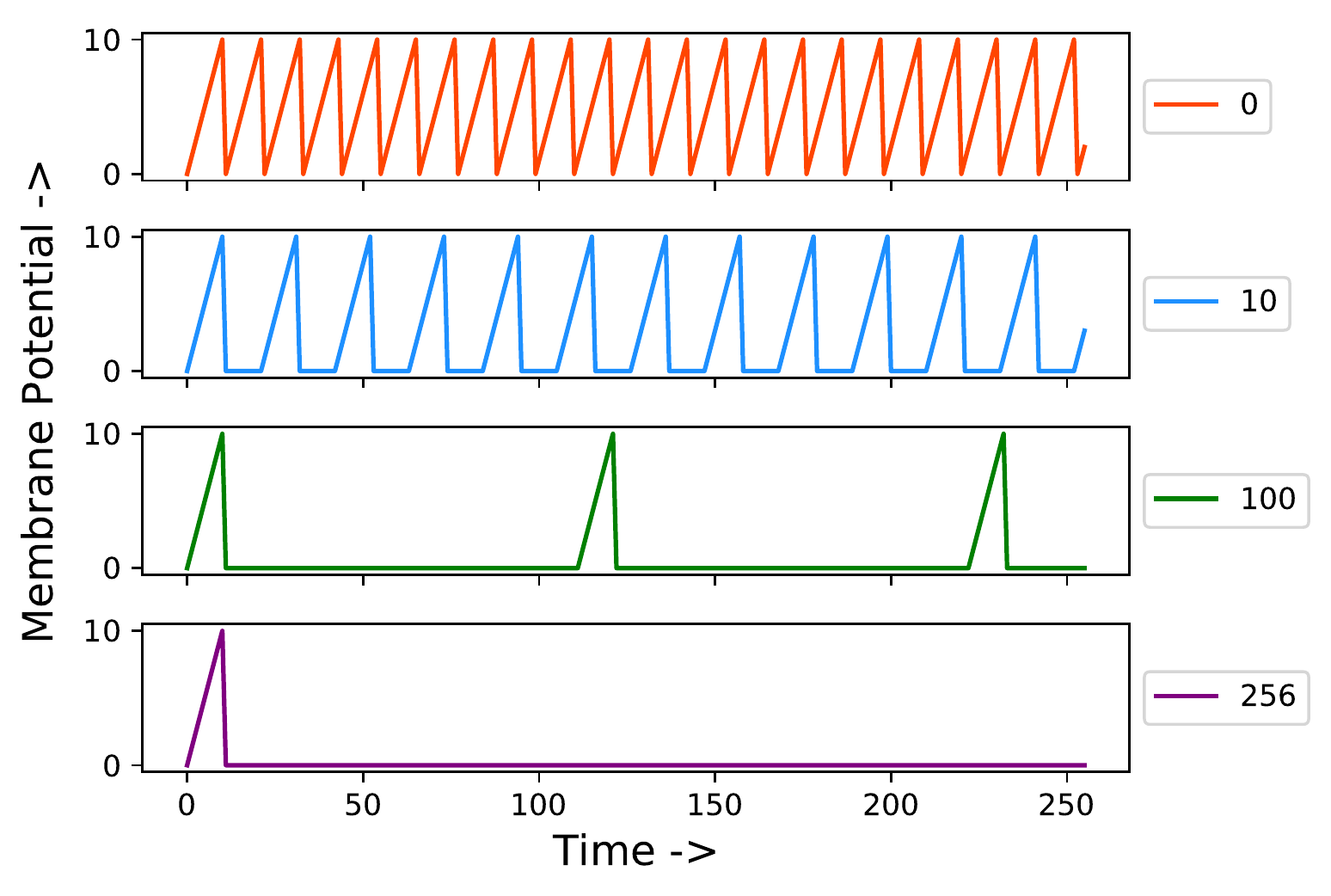} }}%
    \qquad
    \subfloat[\centering]{{\includegraphics[width=0.45\textwidth]{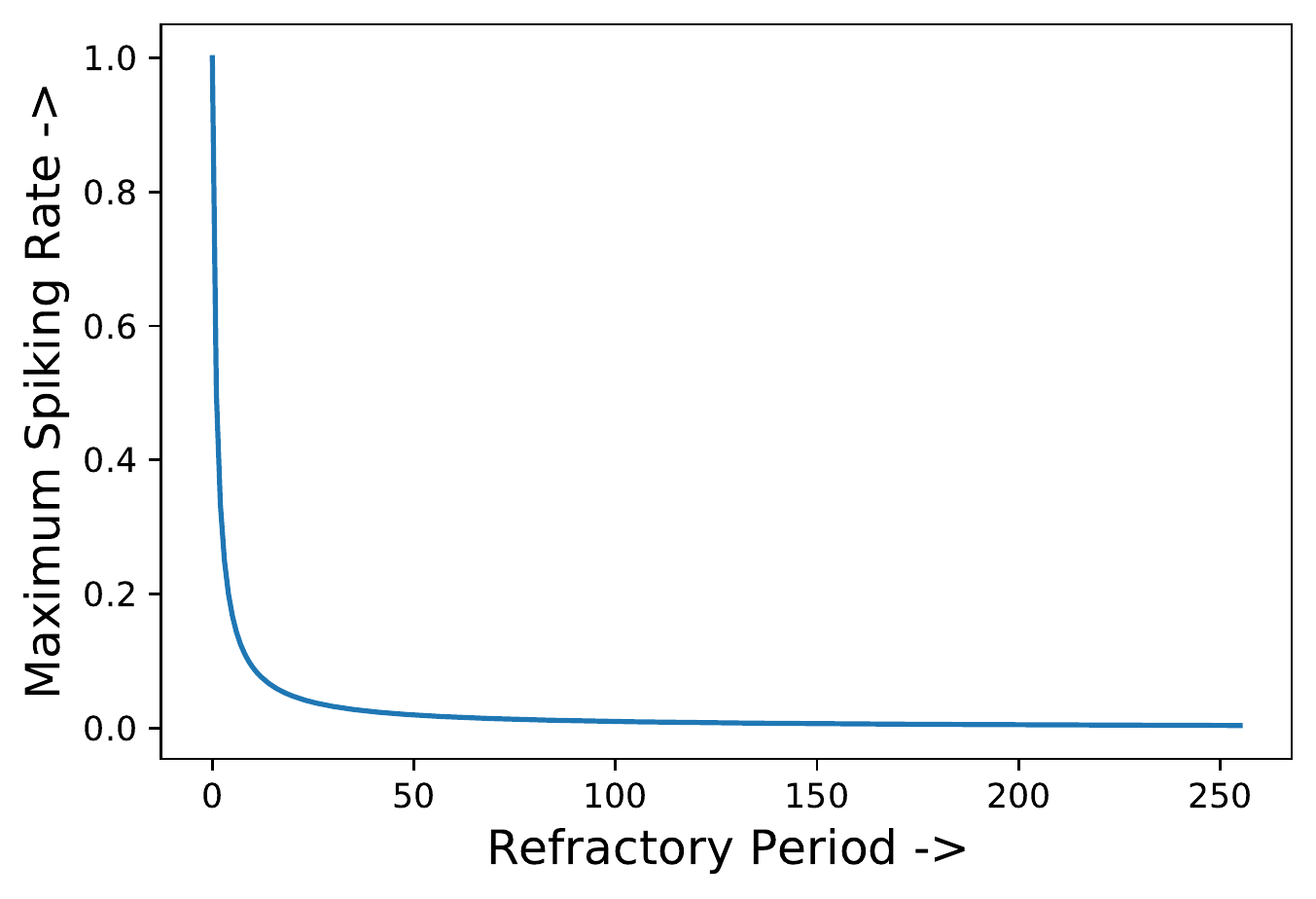} }}%
    \caption{Effect of refractory period on the spiking rate. (a) Membrane Potential of IF neurons having different $t_{ref}$ (0, 10, 100, 256) are shown for a constant unit input ($\theta$=10).  (b) Maximum spiking rate trend with $t_{ref}$ is shown. } 
    \label{refr}
\end{figure}
\begin{equation}
\label{IF}
  V_j^l(t) = \sum\limits_{i} w_{ji}^l \sum\limits_{\tau=1}^{t}S_i^{l-1}(\tau) 
\end{equation}
\noindent if $V_j^l(t) > \theta_j^l$, $V_j^l(t)$ is reset to its resting potential which is taken to be zero and clamped to that voltage for the next $t_{ref}$ time steps. $S_i^{l-1}(t)$ and $w_{ji}^l$ are the input spike train and the input synaptic weights from the $i^{th}$ neuron in the $(l-1)^{th}$ layer.\\ 
\indent Refractory period is the amount of time for which a neuron is unresponsive to stimulus immediately after firing. Higher the refractory period, sparser is the output spike train (see fig.\ref{refr}(a)). It puts an upper bound on the peak spiking rate and as $t_{ref}$ increases, the peak spiking rate falls rapidly (see fig.\ref{refr}(b)). In our network, each neuron has a refractory period  equal to the simulation time ($t_{max}$), which means that any neuron can fire just once throughout the whole simulation, the first time it crosses the threshold $\theta_j^l$ (refer to fig.\ref{refr}(a)).
\begin{equation}
\label{eq:TTFS}
  S_j^l(t) =
  \begin{cases}
  1 &  \text{if} \hspace{2mm}  V_j^l(t) >  \theta_j^l \hspace{2mm} \text{\&} \hspace{2mm} \forall t_0<t, S_j^l(t_0) = 0 \\
    0 & \text{otherwise}
  \end{cases}
\end{equation}
\subsection{Time-to-first-spike Coding}
\begin{figure}
\centering
\includegraphics[width=0.6\textwidth]{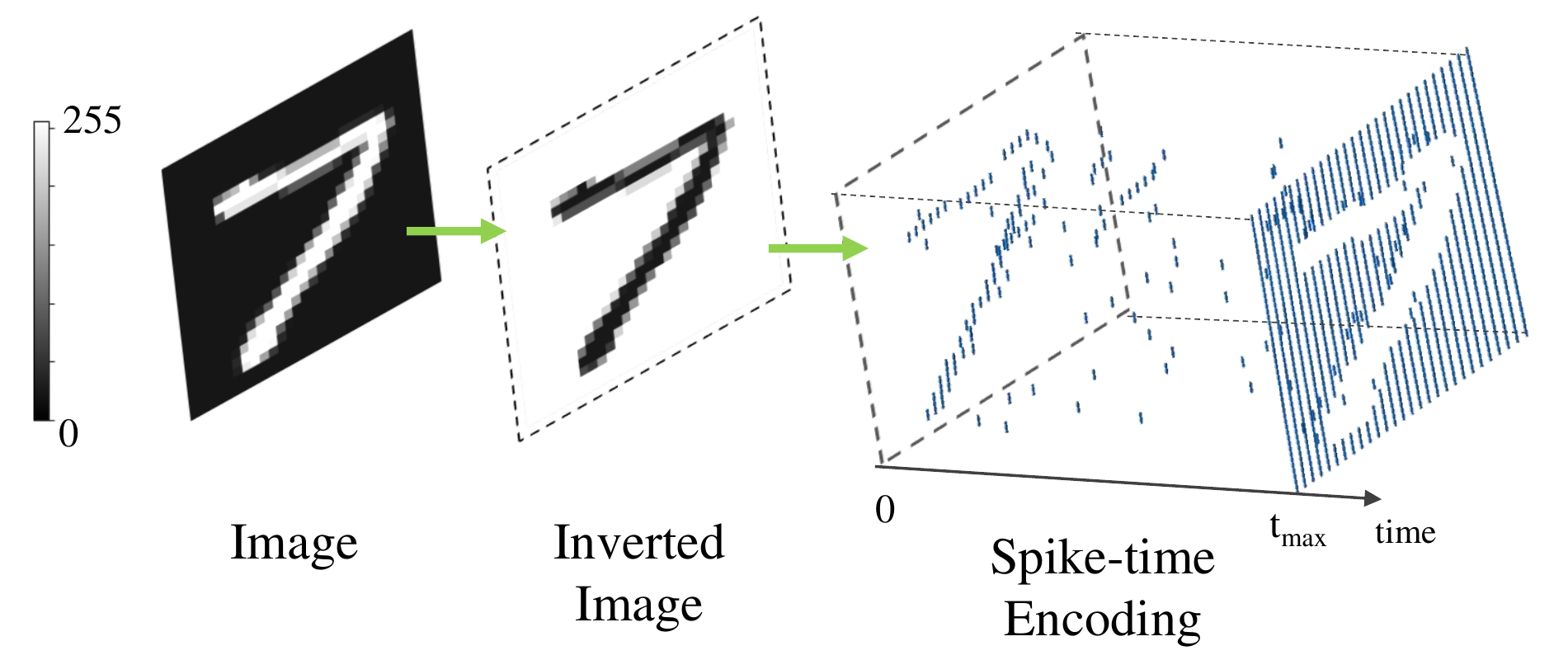}
\caption{Input Spike Encoder: Each input neuron spikes only once. The spike time is set to the intensity of the corresponding pixel in the normalized inverted image scaled by the simulation time (refer to eq.\ref{eq:TTFS}). Larger value of a pixel corresponds to an earlier spike in time.} \label{coding}
\end{figure}
\begin{figure}
\centering
\includegraphics[width=0.6\textwidth]{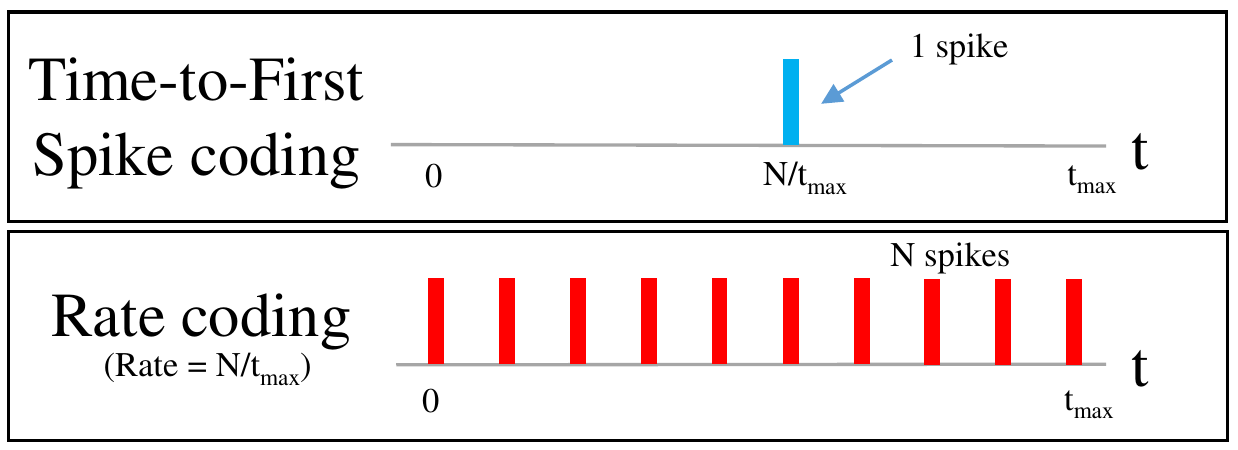}
\caption{Comparing Rate and Time-to-first-spike coding schemes.} \label{rate_TTFS}
\end{figure}
\noindent The input images are encoded to spikes using time-to-first-spike (TTFS) coding. In TTFS coding, the information (here the pixel value) is encoded in the spike time. Consider any channel of the input image with pixel values in the range [0, $I_{max}$].  The spike train $S_j^{in}(t)$ of the $j^{th}$ input neuron is given by :   
\begin{equation}
  S_j^{\hspace{0.4mm} in}(t) =
  \begin{cases}
  1 &  \text{if} \hspace{2mm}  t = \left( \frac{I_{max} - I_j}{I_{max}} \right)t_{max} \\
    0 & \text{otherwise}
  \end{cases}
\end{equation}
where $I_j$ is the $j^{th}$ pixel value. Fig.\ref{coding} demonstrates the input encoding.\\\\
\noindent {\bfseries Comparison with rate coding:} The most common form of neural coding in SNNs is rate coding. In such schemes, the information is encoded in the firing rate of the neuron. This, however, means that the temporal information in the spike trains is lost. In addition to that, the spiking rate has to be controlled to ensure energy efficiency. In problems where the output is expected to be a real number within a range (upto the required precision), controlling the spiking rate is very difficult. For a simulation time of $t_{max}$, for the rate to be equal to N/$t_{max}$, a neuron will have to spike N times. Such an encoding would be highly inefficient and take away almost all of the energy benefits provided by SNNs.\\ \indent
In TTFS coding, only one spike is issued for any case to represent numbers with the same precision, making it highly energy efficient (see fig.\ref{rate_TTFS}). It gets rid of all the extra (redundant) spikes which would be otherwise needed for encoding a particular spike rate. In addition, for classification tasks, inference can be made as soon as the first spike is issued in the output layer. This enables extremely fast information processing in such networks leading to much lower latency.
\subsection{Forward Propagation}
\begin{figure}
\includegraphics[width=\textwidth]{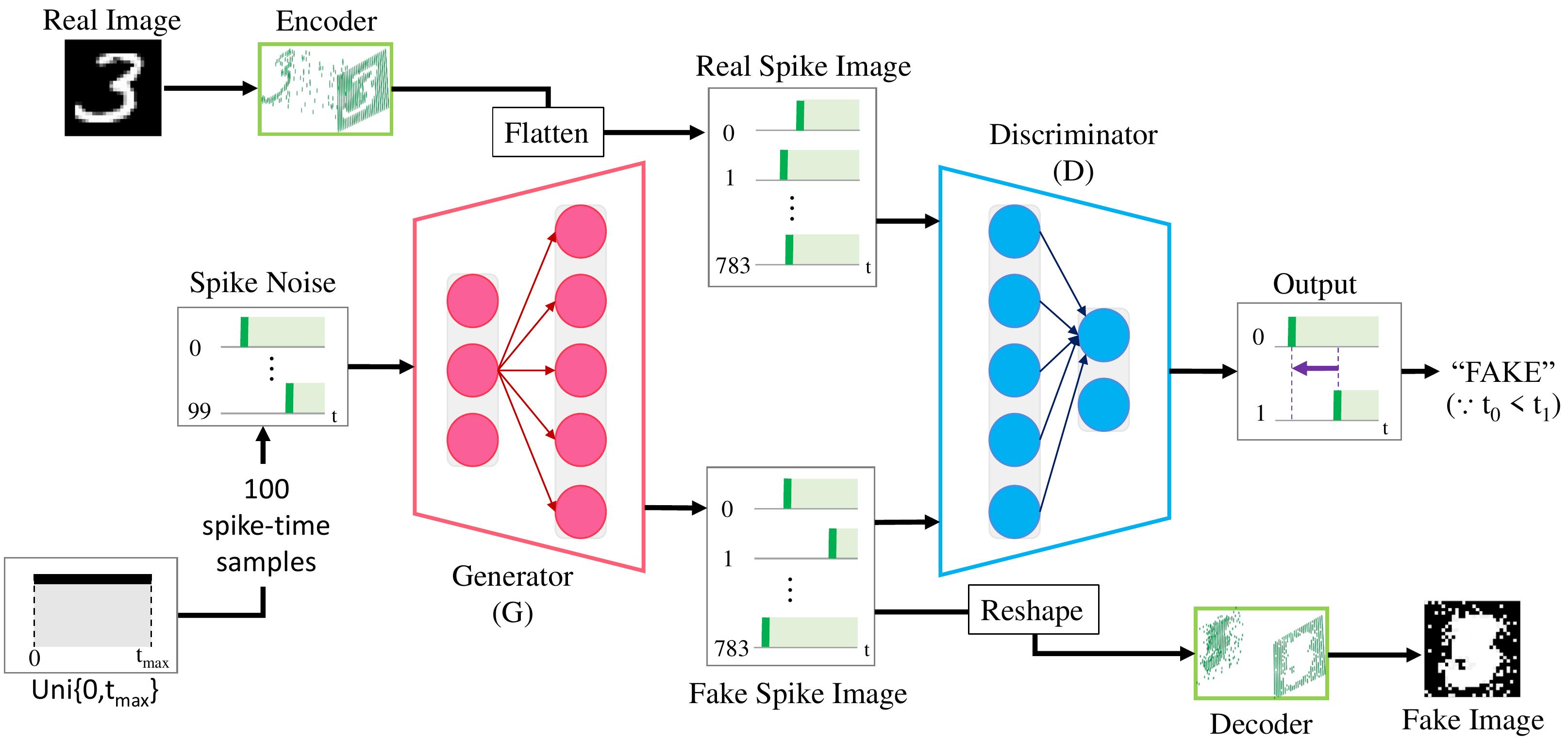}
\caption{Our fully connected (Dense) Spiking-GAN architecture. Input image is converted into a spikemap by the encoder using TTFS coding. Similarly, the input noise vector is converted to a spike-map and given as input to G. Either the real image or the fake image generated from G is given as an input to D. Spikes are propagated through the network and based on the relative spike times of the output neurons in D, the images are labelled as fake or real. } \label{arch}
\end{figure}
Our network architecture consists of the Generator and Discriminator networks, as well as an Encoder and a Decoder for analog-to-spike and spike-to-analog conversion as shown in fig.\ref{arch}.
\\
\noindent {\bfseries Generator:} The generator(G) takes a random noise vector as input. Noise is sampled from a discrete uniform distribution \emph{U\{0, $t_{max}$\}} in our experiments. Similar to the input image, the noise vector is encoded using the time-to-first-spike scheme. Taking this spiking noise(z) as input, the generator outputs a fake spike image $X_{fake} = G(z)$. This image has the same dimensions as the input spike images. The fake spike image is then decoded in an exactly opposite manner to the encoder to produce the fake image.
\begin{figure}[htp]
\centering
    \subfloat[\centering Incorrect class prediction case ($t_0 > t_1$)]{{\includegraphics[width=0.45\textwidth]{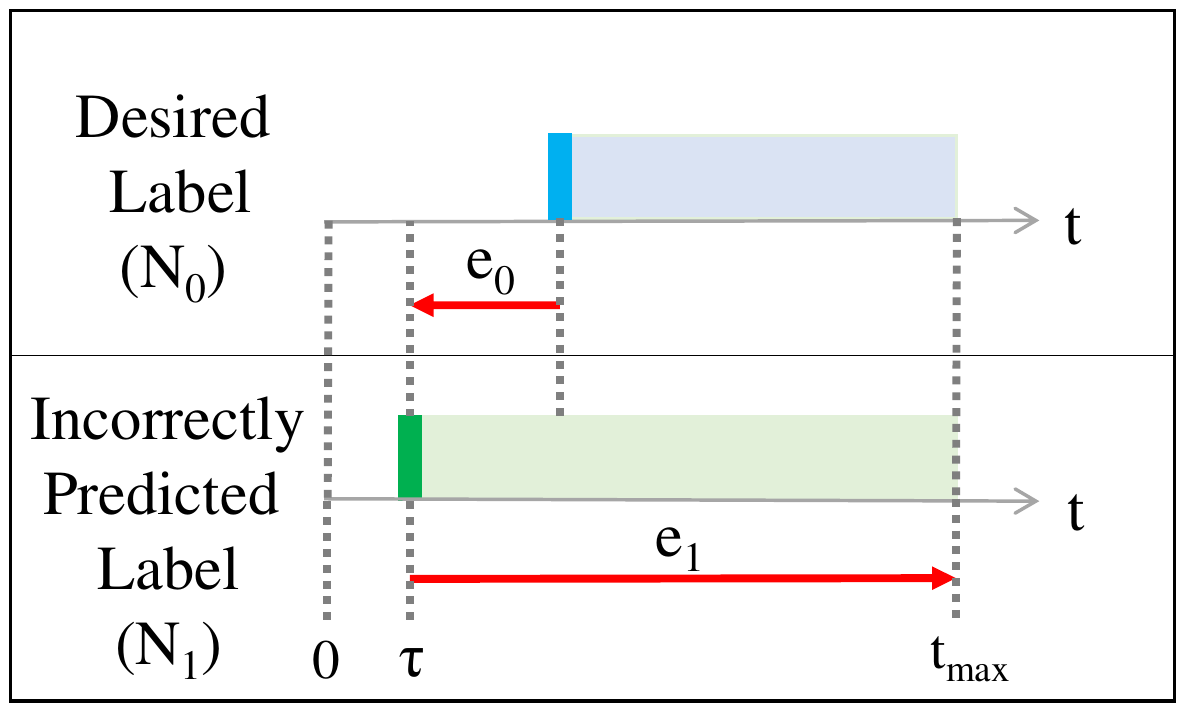} }}%
    \qquad
    \subfloat[\centering Accurate Class Prediction case ($t_0 < t_1$)]{{\includegraphics[width=0.45\textwidth]{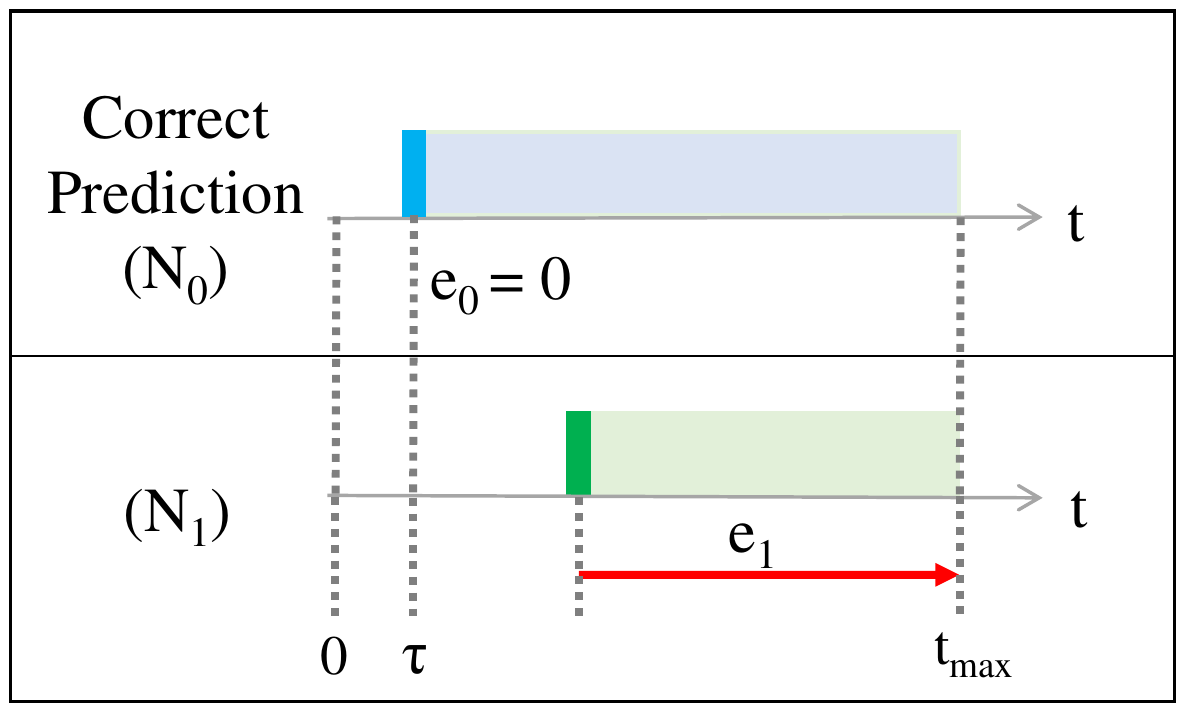} }}%
    \caption{Temporal loss vectors in Spiking-GAN. (when $N_0$ is the desired neuron) (a) $N_1$ incorrectly spikes before $N_0$, so $e_0$ is +ve and $e_1$ is +ve because $N_1$ fires. (b) $N_0$ correctly spikes before $N_1$, so $e_0$ is zero. However, $e_1$ is +ve because $N_1$ fires. $e_1$ is zero only if $N_1$ doesn't fire at all.} 
    \label{loss}
\end{figure}
\\ \noindent {\bfseries Discriminator:} The training images are encoded into the real spike images $X_{real}$ (see fig. \ref{coding}). The discriminator (D) is a binary classifier which takes either the fake spike image ($z \sim P_z(z)$) from the Generator or the real spike image ($x \sim P_{data}(x)$) as input. It has two output neurons: fake (0) and real (1). The class label is determined by the spike timing (see fig.\ref{loss}). If the real neuron spikes before the fake neuron, the input image is classified as real, else fake. 
\subsection{Loss Function}
Various adversarial losses, like minimax loss and Wasserstein loss, have been used for training GANs~\cite{GAN,wgan} but for Spiking-GAN, we have used a least-squares loss~\cite{lsgan} in the temporal domain. We have defined the objective function for Spiking-GAN as follows:
\begin{align}
 \min\limits_{D} \mathcal{L}(D) &= \frac{1}{2}\mathbb{E}_{x \sim P_{data}(x) }\left(\vert\vert D_{t}(x)-T_{d}^{o}\vert\vert^2\right) + \frac{1}{2}\mathbb{E}_{z \sim P_z(z) }\left(\vert\vert D_{t}(G(z))-T_{d}^{o}\vert\vert^2\right) \label{eq:loss1}\\
\min\limits_{G} \mathcal{L}(G) &= \frac{1}{2}\mathbb{E}_{z \sim P_z(z) }\left(\vert\vert D_{t}(G(z))-T_{g}^{o}\vert\vert^2\right)
\label{eq:loss2}\end{align}
\noindent where $D_t$ = $\begin{bmatrix} t_0^o \hspace{0.3mm} & t_1^o \end{bmatrix}$ is a vector of spike times of the output neurons of the discriminator (see fig.\ref{arch}). $T_{d}^o$ = $\begin{bmatrix} T_{d,0}^o \hspace{0.3mm} & T_{d,1}^o \end{bmatrix}$ is the target firing time vector for the discriminator output neurons and $T_{g}^o$ = $\begin{bmatrix} T_{g,0}^o \hspace{0.3mm} & T_{g,1}^o \end{bmatrix}$ is the target firing time vector that the generator wants the discriminator to believe for it's fake data. We use dynamic target firing times that depend on the discriminator output. Let the firing time of the winner output neuron be $\tau$ = $\min(D_t)$ = $\min\{t_0^o, t_1^o\}$. We determine $T_{d}^o$ and $T_{g}^o$ by the following equations: 
  \begin{equation}
  \label{eq:target_d}
        T_{d,1}^o = 
  \begin{cases}
  \tau &  \text{if} \hspace{2mm}  x \sim X_{real} \\
  t_{max}   & \text{if} \hspace{2mm}  x \sim X_{fake}
  \end{cases}
  \hspace{2mm} \text{and} \hspace{2mm}
  T_{d,0}^o = 
  \begin{cases}
  t_{max}  &  \text{if} \hspace{2mm}  x \sim X_{real} \\
  \tau  & \text{if} \hspace{2mm}  x \sim X_{fake}
  \end{cases}
    \end{equation}

\begin{equation}
\label{eq:target_g}
        T_{g,j}^o = 
  \begin{cases}
  \tau  &  \text{if} \hspace{2mm}  j=1 \hspace{7.5mm}\\
  t_{max}  & \text{if} \hspace{2mm}  j=0
  \end{cases}
    \end{equation}
For the desired neuron, the target time is set to $\tau$, so we are training it to emit a spike first. For the other neuron, the target time is set to the maximum possible value, the simulation time $t_{max}$ (refer to fig.\ref{loss}). Note that the target firing time values are flipped for the generator as compared to the discriminator's values for fake images, making the two systems adversaries. 
For the corner case when neither of the output neurons spike, the firing times ($D_t$) are assumed to be $t_{max}$ for loss calculations. Further, $\tau$ in eq.\ref{eq:target_d}-\ref{eq:target_g} is set to 0, thereby heavily penalizing the non-firing desired neuron. This is to incentivize it to fire. We also add an L2 regularization term ($\sum_l\sum_j\sum_i (w_{ij}^l)^2$) in the loss function.
\subsection{Temporal Backpropagation}
For training the network we use stochastic gradient descent. We update $w_{ji}^l$, the weight for the connection from the $i^{th}$ neuron of the $(l-1)^{th}$ layer to the $j^{th}$ neuron of the $l^{th}$ layer as follows: ($\eta$ is the learning rate)
\begin{equation}
        w_{ji}^l = w_{ji}^l -  \eta\frac{\partial \mathcal{L}}{\partial w_{ji}^l}
    \end{equation}
Where appropriate $\mathcal{L}$ from eq.\ref{eq:loss1}-\ref{eq:loss2} is chosen for training the respective network. But calculating these weight updates is a problem. The IF neuron is not a differentiable activation function. However, it approximates ReLU~\cite{bp-stdp, s4nn}. For a neuron which has a ReLU activation function, the output of the $j^{th}$ neuron of the $l^{th}$ layer $y_j^l$ is given by:
\begin{equation}
        y_j^l = \max\left(0, z_j^l = \sum\limits_i w_{ji}^lx_i^{l-1}\right)
        \label{relu}
    \end{equation}
where $x_i^{l-1}$ is the input from the $i^{th}$ neuron in the previous layer and $w_{ji}^l$ is the corresponding input weight. $z_j^l$ for ReLU is equivalent to $V_j^l$ in our model for a given time step. For identical weights, larger input values correspond to a larger value of $y_j^l$. Similarly, in the TTFS model where all the information is encoded in the spike time, larger the input, higher is the membrane potential and earlier is the spike. So, we assume an equivalence relationship between the firing time of the IF neuron $t_j^l$ and the corresponding output of a ReLU neuron $y_j^l$:
\begin{equation}
\label{eq:approx}
    y_j^l \sim t_{max} - t_j^l
    \end{equation}
Each neuron only spikes once. So, for any neuron only the pre-synaptic inputs that fire before it contribute to its spike time. So, using eq.\ref{eq:approx}, we further assume: 
\begin{equation}
\label{deriv}
            \frac{\partial t_j^l }{\partial V_j^l}  = 
  \begin{cases}
  -1  &  \text{if} \hspace{2mm}  t_j^l < t_{max}\\ 
  0  & \text{otherwise}
  \end{cases}
    \end{equation}
Based on eq.\ref{IF} \& eq.\ref{relu}-\ref{deriv}, the derivatives for ReLU and IF respectively would be:
\begin{equation}
        \frac{\partial y_j^l}{\partial w_{ji}^l} = \frac{\partial y_j^l}{\partial z_j^l} \frac{\partial z_j^l}{\partial w_{ji}^l} = \begin{cases}
  x_i^{l-1}  &  \text{if} \hspace{2mm}  y_j^l>0\\
  0  & \text{otherwise}
  \end{cases}
        \label{drelu}
    \end{equation}
\begin{equation}
        \frac{\partial t_j^l }{\partial w_{ji}^l} = \frac{\partial t_j^l }{\partial V_j^l} \frac{\partial V_j^l}{\partial w_{ji}^l} = 
  \begin{cases}
  -\sum\limits_{\tau=1}^{t_j^l} S_i^{l-1}(\tau)  &  \text{if} \hspace{2mm}  t_j^l < t_{max}\\ 
  0  & \text{otherwise}
  \end{cases}
    \end{equation}
Which in turn gives us the loss gradient:
\begin{equation}
            \frac{\partial \mathcal{L} }{\partial w_{ji}^l}  = \frac{\partial \mathcal{L} }{\partial t_j^l} \frac{\partial t_j^l}{\partial w_{ji}^l} = 
  \begin{cases}
  -\delta_j^l\sum\limits_{\tau=1}^{t_j^l} S_i^{l-1}(\tau)  &  \text{if} \hspace{2mm}  t_j^l < t_{max}\\ 
  0  & \text{otherwise}
  \end{cases}
    \end{equation}
where $\delta_j^l = \partial \mathcal{L}/\partial t_{j}^l$. To calculate $\delta_j^l$, we backpropagate the gradient (using chain rule) for the hidden layers:
\begin{align}
     \delta_j^l &= \partial \mathcal{L}/\partial t_{j}^l =  \mathlarger{\mathlarger{\sum}}\limits_k\left( \frac{\partial \mathcal{L}}{\partial t_k^{l+1}}\frac{\partial t_k^{l+1}}{\partial V_k^{l+1}}\frac{\partial V_k^{l+1}}{\partial t_j^{l}}\right)\nonumber\\ &=
     \begin{cases}
  \sum\limits_k \delta_k^{l+1}w_{kj}^{l+1} &  \text{if} \hspace{2mm}  t_j^l < t_k^{l+1} \\
  0  & \text{otherwise}
  \end{cases}
\end{align}
For the output layer, $\delta_j^{o}$ is given by:
\begin{align}
    \delta_j^{o}(D) =  \frac{\partial \mathcal{L}(D)}{\partial t_j^{o}} &= \left[ D_t(x) - T_d^o \right]_{x\sim P_{data}(x)} + \left[ D_t(G(z)) - T_d^o \right]_{z\sim P_z(z)} \\
    \delta_j^{o}(G) =  \frac{\partial \mathcal{L}(G)}{\partial t_j^{o}} &= \left[ D_t(G(z)) - T_g^o \right]_{z\sim P_z(z)} 
\end{align}
At every layer, we normalize the values of $\delta_j^l$ by dividing it by its L1 norm ($\sum_j\delta_j^l$) prior to calculating the weight updates.
\subsection{Training}
The Discriminator (D) and Generator (G) are trained together in an alternating manner. While training D, the fake images generated by G are used but weights of G are not altered. Similarly, while training G, the prediction of D is used and the loss is backpropagated through D to G but the weights of D are kept unchanged. We train both G and D once every epoch. Once trained, the generator G which now (approximately) samples from the same distribution as the training dataset is used to create fake samples.
\section{Results}
\subsection{Classification}
To demonstrate the effectiveness of the `Aggressive TTFS' loss function, we trained a 2-layer fully-connected 784-400-10 network to classify MNIST digits.
Our modified objective forces most of the `incorrect' neurons to not spike at all and heavily penalizes the actual label neuron in case it doesn't spike
\begin{figure}
\centering
    \subfloat[\centering Mean inference time comparison]{{\includegraphics[width=0.45\textwidth]{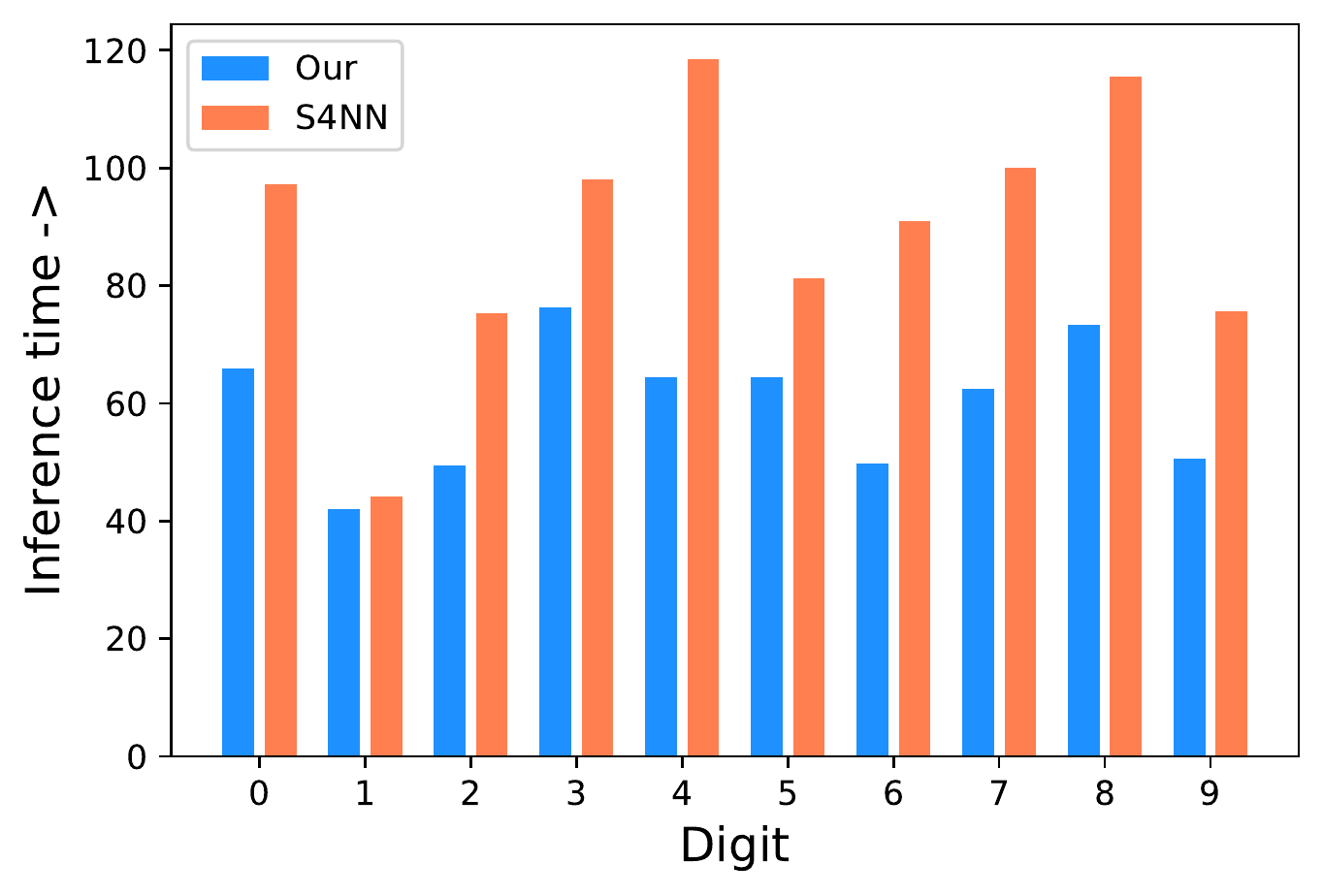} }}%
    \qquad
    \subfloat[\centering Mean spike count comparison]{{\includegraphics[width=0.45\textwidth]{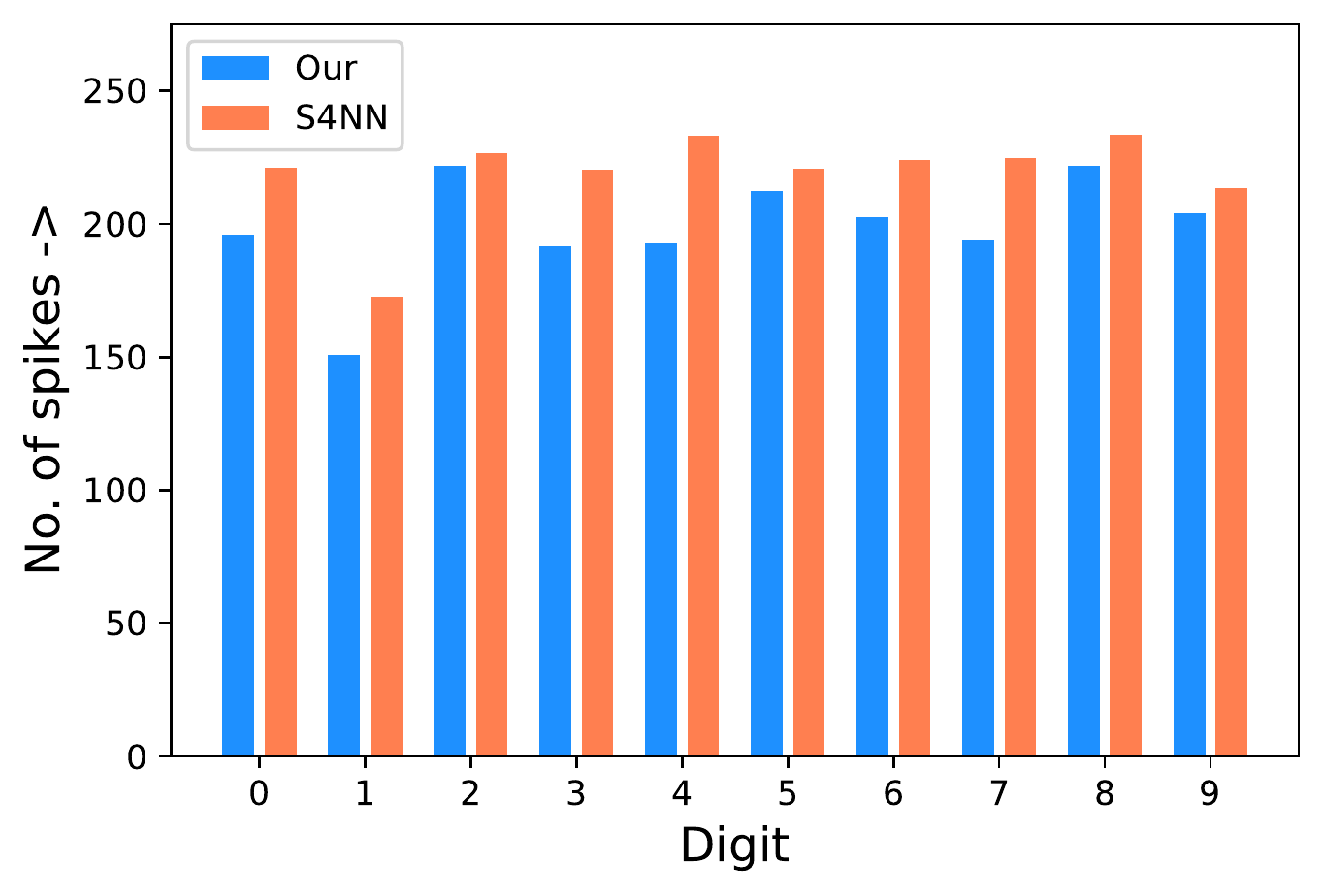} }}%
    \caption{Comparison of the mean inference time and mean spike count in the whole network needed for classification of each digit (with S4NN) using a 2-layer dense (784-400-10) network. Our loss function yields better performance for every digit.}
    \label{comp}
\end{figure}
 \begin{figure}[htp]
 \vspace{-2mm}
\centering
    \subfloat[\centering S4NN learned filters]{{\includegraphics[width=0.45\textwidth]{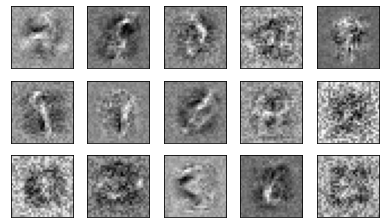} }}%
    \qquad
    \subfloat[\centering Aggressive TTFS learned filters]{{\includegraphics[width=0.45\textwidth]{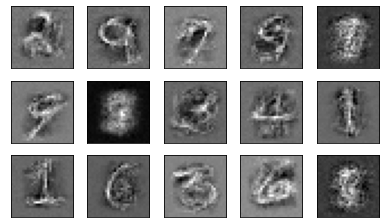} }}%
    \caption{Comparison between 15 randomly picked filters from the 400 (28x28) hidden weights of a 784-400-10 dense network trained for MNIST classification. Filters learned using our method are significantly less noisy.} 
    \label{comp_vis}
\end{figure}
(refer to sec.2.4).
This significantly reduces the inference latency and increases the sparsity of the network as compared to the relative margin-based objective in S4NN which imposes only a small penalty on the `incorrect' neurons and that too only when they fire very close to the ground truth neuron. It also has a much lower penalty when the desired neuron doesn't fire at all.
\\ \indent
Fig.\ref{comp}(a),(b) show the digit-by-digit comparisons for the same. Our network, on an average, takes 59.7 time steps and needs only 194 spikes to make a decision on the class label, which is 33.4\% faster inference in 11.1\% fewer spikes (S4NN needs 218.3 spikes and takes 89.7 time steps on an average). So, on an average only 16.2\% of the neurons in the network (1194) fire and the network makes its decision in 23.3\% of the simulation time (256). Note that the network decision can be taken as soon as any of the output neuron spikes and all the spike counts are the average number of spikes issued in the whole network till that time step. Fig.\ref{comp_vis}(a),(b) compares the filters learned by the two methods. 15 randomly chosen (28x28) weights from the hidden layer are shown. Aggressive TTFS results in qualitatively superior filters, they are much sharper and less noisy.  However, the increased sparsity and faster inference lead to a slightly lower test accuracy of 96.7\% (as compared to 97.4\%). Though with a bigger network 784-1000-10, we were able to achieve a test accuracy of 97.6\%, which surpasses the classification performance of other temporal coding-based works that use simple instantaneous synaptic current kernels. 
\subsection{Spiking-GAN}
\begin{figure}
\centering
    \subfloat[\centering ][Images from the MNIST dataset (training)]{{\includegraphics[width=0.8\textwidth]{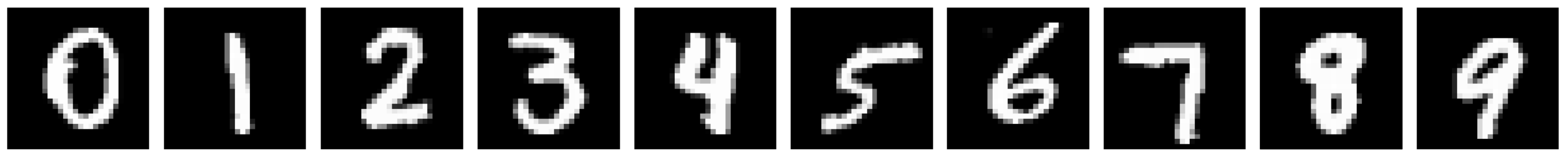} }}%
    \qquad
    \subfloat[\centering ][Selected images generated by an ANN-based GAN]{{\includegraphics[width=0.8\textwidth]{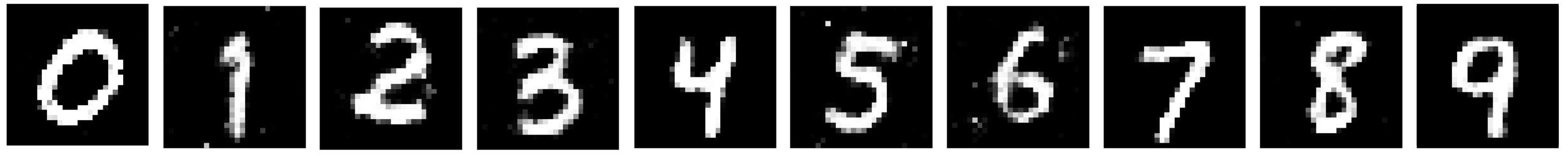} }}%
    \qquad
    \subfloat[\centering ][Selected images generated by the Spiking-GAN]{{\includegraphics[width=0.8\textwidth]{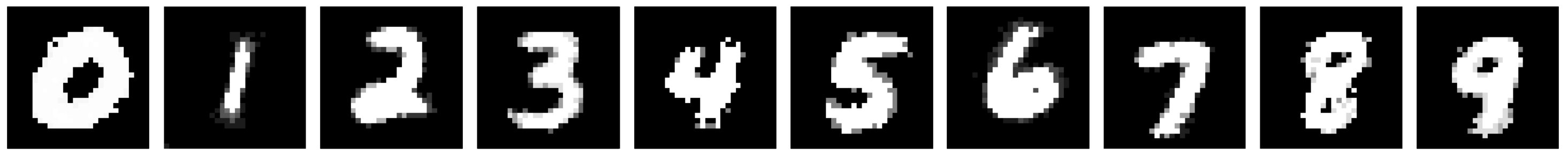} }}%
    \caption{Comparison between the training images, some of the good individual digit samples generated by the Spiking-GAN and some of the selected samples from an equivalent ANN-based GAN.}
    \label{results}
\end{figure}

 We trained our network on the MNIST dataset for each individual digit. The generator is a 2-layer fully connected (dense) network (100-400-784). 100 spike noise images are given as an input and the flattened generated fake sample is the output. The discriminator is a 2-layer fully connected (dense) network (784-400-2) which takes a flattened spike train as input. Fig.\ref{results}(c) shows some selected samples generated by G after training the network for 50 epochs. As can be seen the samples generated are of high quality.
\\ \indent Fig.\ref{results}(b) shows the samples generated by an equivalent ANN-based GAN with a 100-400-784 generator and 784-400-1 discriminator. The output neurons of G use tanh activation function. The output neuron of D has a sigmoid activation function and consequently only one neuron is needed for classification. Rest of the neurons are normal ReLU neurons. As can be seen from fig.\ref{results}, the quality of samples generated by Spiking-GAN is comparable to the ANN-based GAN.
 \\ \indent The computational costs of SNNs is proportional to the number of spikes issued in the network~\cite{J_Kim}. 
 Each additional spike in a neuron leads to an additional operation in each post synaptic neuron~\cite{eth18}. 
 The Spiking-GAN network has extremely sparse spike trains. Since all the neurons spike at most once, the theoretical maximum number of spikes in the complete network is equal to the number of neurons in the network (2470 in our case).
 \\\indent To better demonstrate the sparsity and efficiency of the network, consider the MNIST dataset. Let the total simulation be of 256 time-steps.  For a simple lossless spike-rate/spike-count based coding scheme where the number of spikes denote the input value, the mean number of spikes needed for just the input coding is 173,745 per image (or 222 spikes per pixel) for an MNIST image. As large number of values for any given image is large, for efficient coding we would invert the input image. 
 Even for an inverted image, the mean number of spikes needed for the input is 26,175 per image (or 33 spikes per pixel). 
 For TTFS coding this number reduces to 784 spikes per image (or 1 spike per pixel) which is a 33x improvement in the input coding efficiency. The lower number of spikes show efficient information transfer and in turn enables lower energy requirements when implemented on neuromorphic hardware.
  \begin{figure}
\centering
    \subfloat[\centering Generated image evolution ]{{\includegraphics[width=0.37\textwidth]{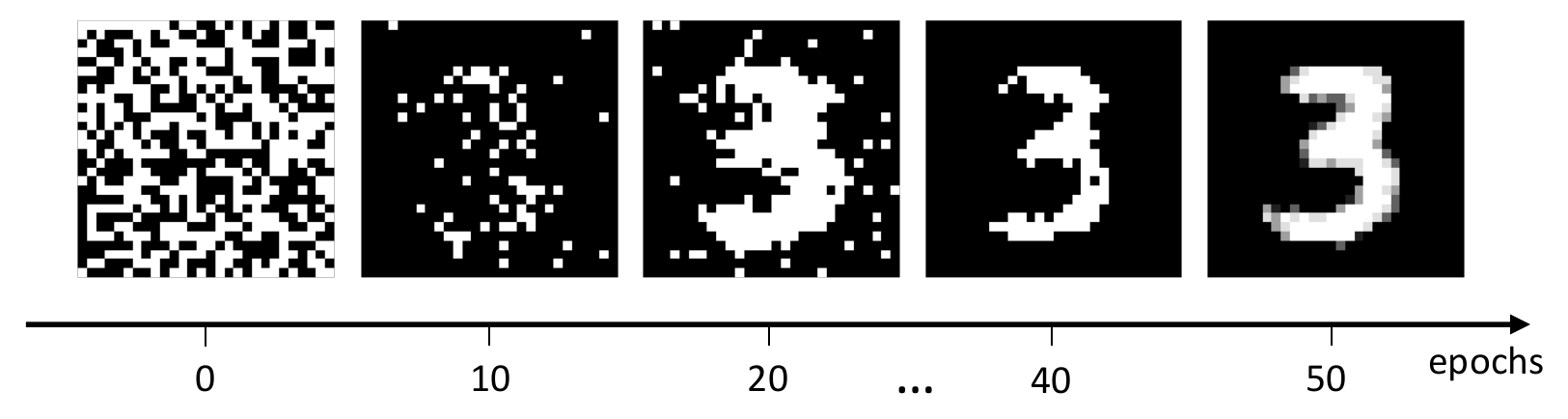} }}%
    \qquad
    \subfloat[\centering Some `undesired' samples]{{\includegraphics[width=0.5\textwidth]{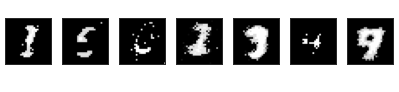} }}%
    \caption{(a) Trend of generated image quality with training epochs. (b) Selected interpretable `undesired' samples generated by a trained Spiking-GAN. } 
    \label{training}
\end{figure} 
\section{Conclusion}
In this paper, we have demonstrated the viability of implementing a generative adversarial network in the spiking domain. The network encodes and transfer information in a highly sparse manner using TTFS coding. We trained the Spiking-GAN using spike timing-based learning rules. Such a framework could be easily adapted and extended to realize other generative networks like variational autoencoders (VAEs), conditional-GANs and auxiliary classifier-GANs in the spiking domain. More importantly, it can prove very useful for solving other regression problems like image translation, image segmentation and even combined classification-regression tasks like object detection in the spiking domain using spike-based learning rules. At the same time, ensuring the sparsity, energy efficiency and low latency while realizing such networks.   
\bibliographystyle{splncs03_unsrt}
\bibliography{ref.bib}
\end{document}